\documentclass[runningheads]{llncs}
\usepackage{graphicx}

\usepackage{tikz}
\usepackage{comment}
\usepackage{amsmath,amssymb} 
\usepackage{color}
\usepackage[pagebackref,breaklinks,colorlinks]{hyperref}
\usepackage{subfigure}
\usepackage{booktabs}
\usepackage{bm}
\usepackage{xcolor,colortbl}
\usepackage{bbm}
\usepackage{multirow}
\usepackage{wrapfig}
\usepackage{mathtools}
\usepackage{algorithm}
\usepackage{listings}

\definecolor{amber}{rgb}{1.0, 0.75, 0.0}
\definecolor{cyan}{cmyk}{.3,0,0,0}
\usepackage{soul}
\definecolor{airforceblue}{rgb}{0.36, 0.54, 0.66}

\usepackage[accsupp]{axessibility}  
\usepackage{cite}

\begin{document}
\pagestyle{headings}
\mainmatter
\def\ECCVSubNumber{2219}  

\title{Iwin: Human-Object Interaction Detection via Transformer with Irregular Windows} 

\titlerunning{Iwin}
%
\author{Danyang Tu\inst{1} \and
Xiongkuo Min\inst{1} \and
Huiyu Duan\inst{1} \and
Guodong Guo\inst{2} \and \\
Guangtao Zhai\inst{1}  \and
Wei Shen \inst{3} }
\authorrunning{Danyang Tu et al.}
%
\institute{Institute of Image Communication and Network Engineering, Shanghai Jiao Tong University \and
           Institute of Deep Learning, Baidu Research, Beijing, China\\ \and
           MoE Key Lab of Artificial Intelligence, AI Institute, Shanghai Jiao Tong University
\email{\{danyangtu, minxiongkuo, huiyuduan, zhaiguangtao, wei.shen\}@sjtu.edu.cn, guoguodong01@baidu.com}}
\maketitle

\begin{abstract}

This paper presents a new vision Transformer, named Iwin Transformer, which is specifically designed for human-object interaction (HOI) detection, a detailed scene understanding task involving a sequential process of human/object detection and interaction recognition. Iwin Transformer is a hierarchical Transformer which progressively performs token representation learning and token agglomeration within \underline{\textbf{i}}rregular \underline{\textbf{win}}dows. The irregular windows, achieved by augmenting regular grid locations with learned offsets, 1) eliminate redundancy in token representation learning, which leads to efficient human/object detection, and 2) enable the agglomerated tokens to align with humans/objects with different shapes, which facilitates the acquisition of highly-abstracted visual semantics for interaction recognition. The effectiveness and efficiency of Iwin Transformer are verified on the two standard HOI detection benchmark datasets, HICO-DET and V-COCO. Results show our method outperforms existing Transformers-based methods by large margins (3.7 mAP gain on HICO-DET and 2.0 mAP gain on V-COCO) with fewer training epochs ($0.5 \times$).
\end{abstract}

\section{Introduction}

Given an image containing several humans and objects, the goal of human-object interaction (HOI) detection is to localize each pair of human and object as well as to recognize their interaction. It has attracted considerable research interests recently for its great potential in the high-level human-centric scene understanding tasks.

\begin{figure}[t]

    		\centering
    		\includegraphics[width=0.98\linewidth]{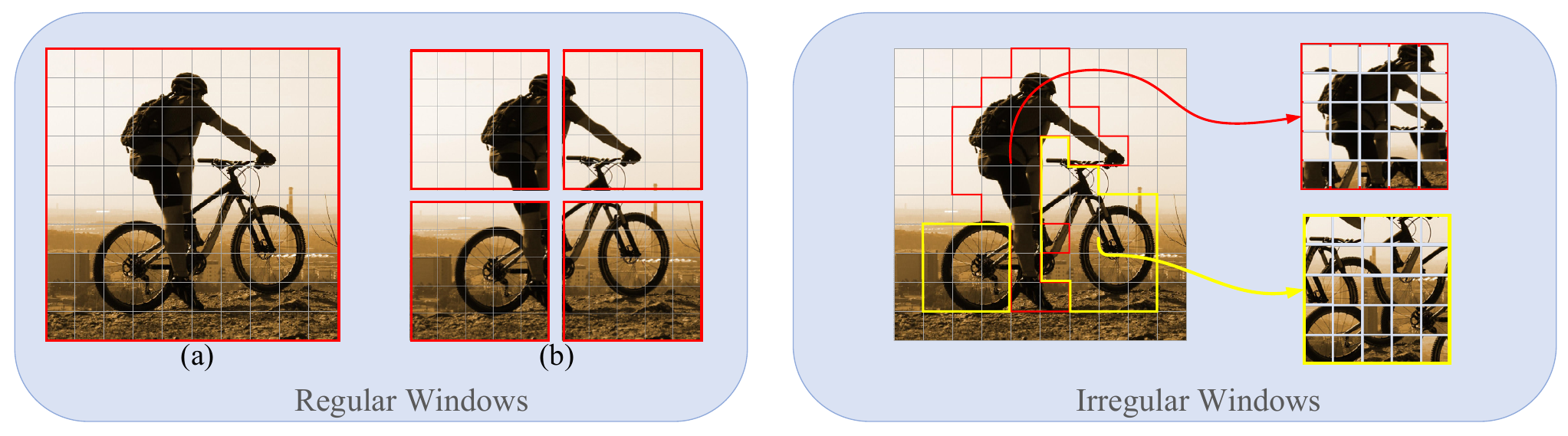}
    		\caption{ An illustration of {\bfseries regular windows} and {\bfseries irregular windows}. For regular window, the window can be a global rectangle containing the entire image (\textbf{a}) or a local one (\textbf{b}) that might divide an object into parts. In contrast, irregular windows can align with the arbitrary shapes of objects or semantic regions. We can rearrange the tokens within an irregular window to form a rectangular window for self-attention computation.} 
    		\label{ill}

    \end{figure}

Recently, vision Transformers~\cite{attention} have started to revolutionize the HOI detection task, which enable end-to-end HOI detection and achieve leading performance~\cite{qpic,endd,hotr,chen2021reformulating,zhang2021mining, tu2022video}. However, they mainly adopt DETR-like~\cite{detr} Transformers, which learn visual semantics by self-attention between patch tokens within rectangular windows (Figure.~\ref{ill}), either local or global, resulting in mixed and redundant visual semantics. Such visual semantics fail to capture object-level abstractions, which are crucial for interaction recognition.

    \begin{figure}[t]
    \centering
    \begin{minipage}[t]{0.49\textwidth}
    \centering
    \includegraphics[width=0.99\linewidth]{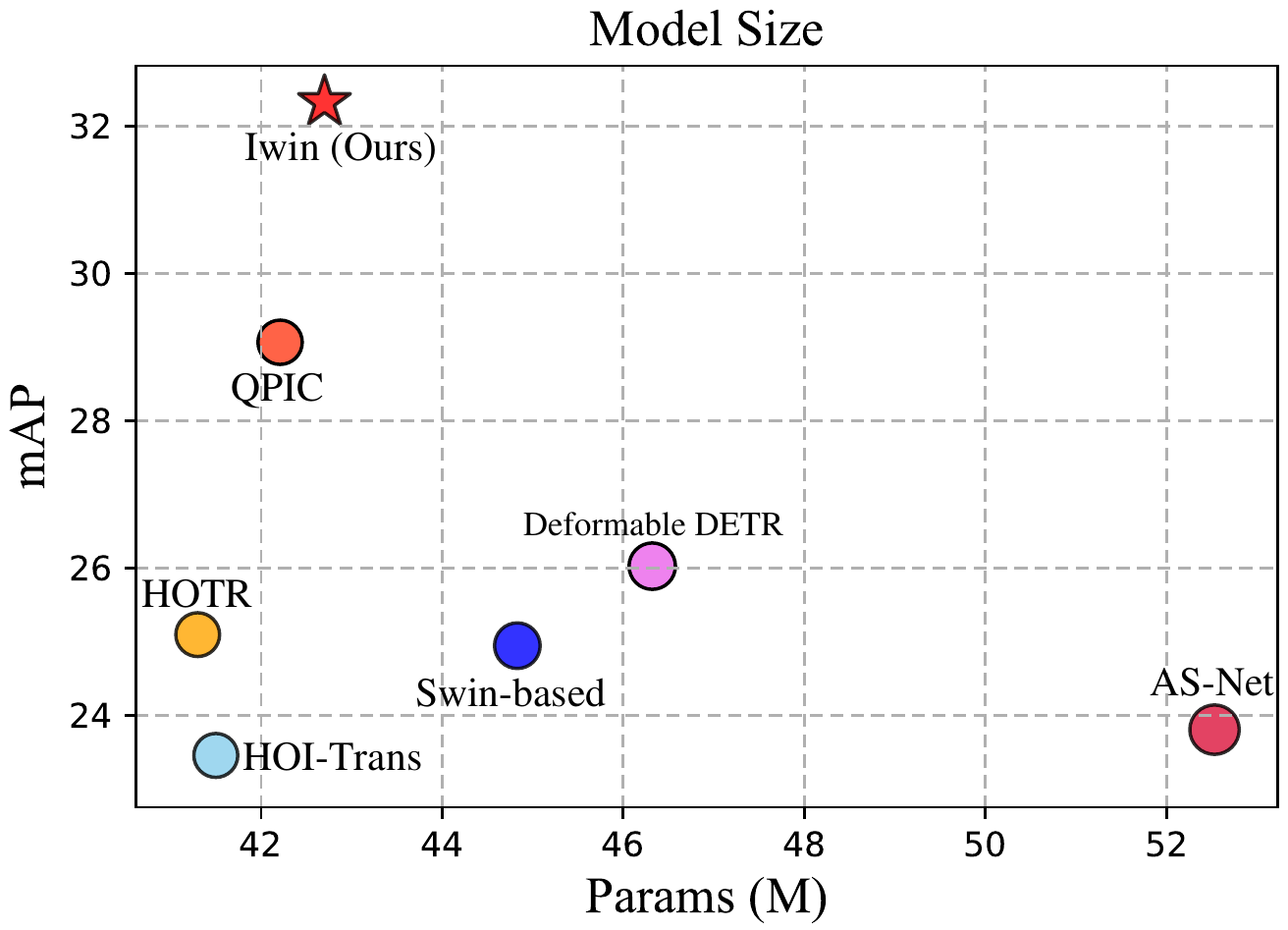}
    \end{minipage}
    \begin{minipage}[t]{0.49\textwidth}
    \centering
    \includegraphics[width=0.99\linewidth]{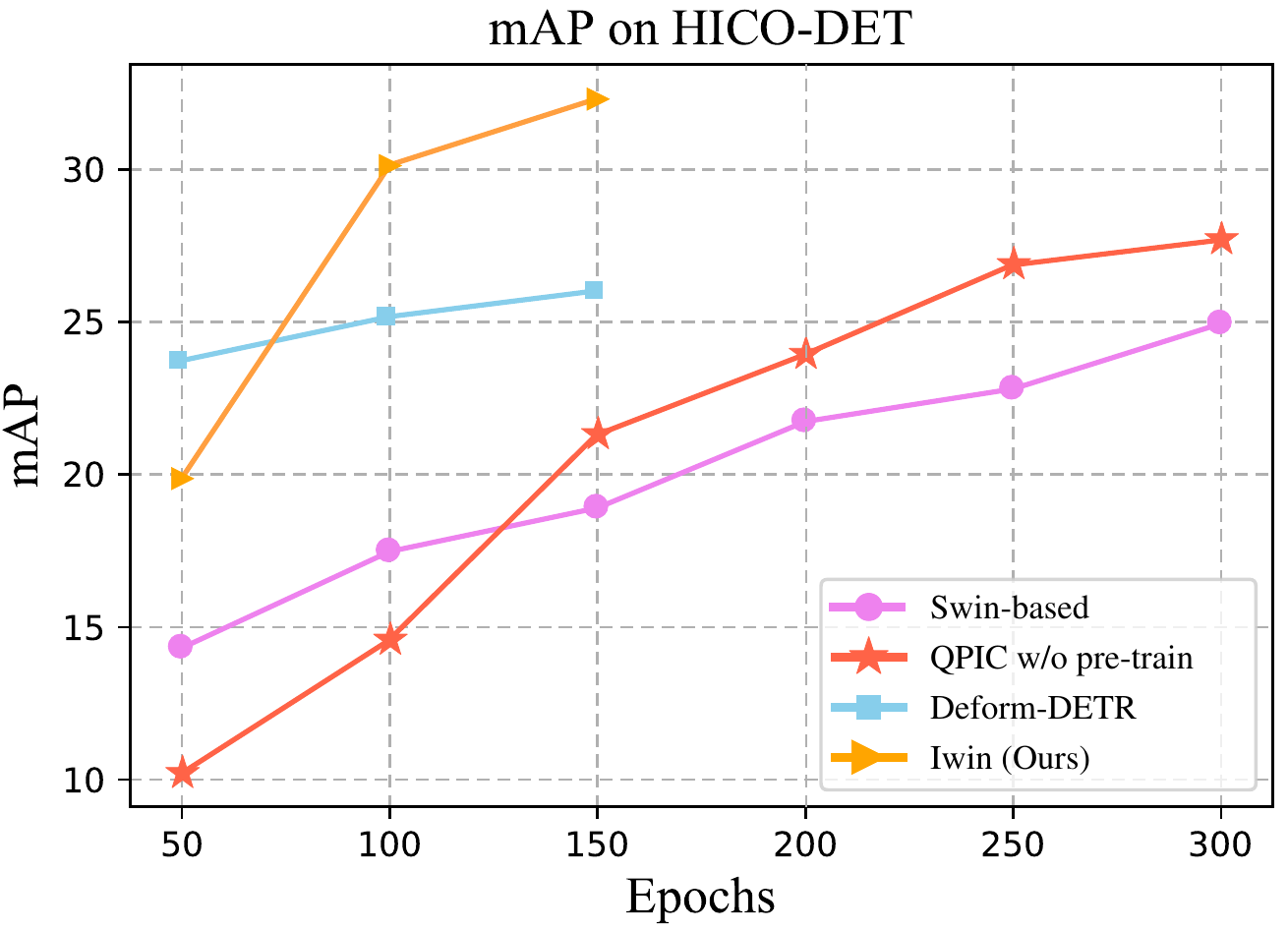}
    \end{minipage}
    \caption{\textbf{Model size} and \textbf{training epochs} vs. performance analysis for HOI detection on HICO-DET. Specifically, Swin-based refers to perform a Swin-T model on the output of a FPN as the encoder of a Transformer. For fair comparison, all the models in the right sub-figure are trained from scratch.}

    \label{first}
    \end{figure}

To address the aforementioned limitation, we propose Iwin Transformer, short for Transformer with \underline{\textbf{i}}rregular \underline{\textbf{win}}dows, which is a new hierarchical vision Transformer specifically designed for HOI detection. The irregular windows, as illustrated in Figure~\ref{ill}, are obtained by augmenting regular grid locations with learned offsets, which are expected to be aligned with humans/objects with arbitrary shapes. Iwin Transformer performs both token representation learning and token agglomeration within irregular windows. The former eliminates redundancy in self-attention computation between patch-level tokens, leading to efficient human/object detection; The latter progressively structurizes an image as a few agglomerated tokens with highly-abstracted visual semantics, the contextual relations between which can be easily captured for interaction recognition. Iwin Transformer takes the characteristic of HOI detection into account, \emph{i.e.}, a sequential process of human/object detection and interaction recognition, enjoying both higher efficiency and better effectiveness, as shown in Figure~\ref{first}.

It is worth emphasizing that our method is different from regular window-based Transformers as well as deformable DETR \cite{deformableDETR}. In addition to the different objectives,  the former,\emph{e.g.}, Swin Transformer \cite{swin}, partitions an image into several regular windows that are not aligned with humans/objects with different shapes, leading to redundancy in self-attention computation; The latter, \emph{i.e.}, deformable DETR, deals with a fixed number of tokens without token grouping, being weak in extracting highly-abstracted visual semantics.

Experimental results show that Iwin Transformer outperforms existing SOTA methods by large margins and is much easier to train. Specifically, Iwin achieves a 3.7 mAP gain on HICO-DET~\cite{chao2018learning} and a 2.0 mAP gain on V-COCO~\cite{gupta2015visual} with fewer training epochs ($0.5 \times $).

\section{Related Work}
{\bfseries CNN-Based HOI Detection}. CNN-based methods can be divided into two different types:1) Two-stage. Most two-stage methods~\cite{chao2018learning,drg,ican,interactions,no-frills,zhi_eccv2020,uniondet,kim2020detecting,li2020detailed,li2019transferable,lin2020action,liu2020amplifying,qi2018learning,ulutan2020vsgnet,wan2019pose,wang2020contextual,wang2019deep,xu2019interact,yang2020graph,zhong2021polysemy,zhou2019relation,zhou2020cascaded} obtain the human and object proposals by using a pre-trained object detector firstly, and then predict interaction labels by combining features from localized regions. More specifically, they first use Faster R-CNN \cite{ren2015faster} or Mask R-CNN \cite{he2017mask} to localize targets, including humans and objects. Then, the cropped features are fed into a multi-stream network, which normally contains a human stream, an object stream and a pairwise stream. The first two process features of target humans and objects, respectively, and the last one normally processes some auxiliary features, such as spatial configurations of the targets, human poses, gaze~\cite{tu2022end}, or the combination of them. In addition, some other methods utilize graph neural networks to refine the features \cite{qi2018learning,ulutan2020vsgnet,wang2020contextual,yang2020graph,zhou2019relation,zhang2021spatially}. Such methods have made impressive progress in HOI detection, but they still suffer from low efficiency and effectiveness. 2) Single-stage. Different from two-stage methods, single-stage methods detect the targets and their interactions simultaneously. In detail, UnionDet \cite{uniondet} predicts the interactions for each human-object union box by an anchor-based method. PPDM \cite{ppdm} and IPNet \cite{ipnet} represent the interactions as the midpoints of human-object pairs and detect them based on point detection networks. GGNet \cite{ggnet} further improves the performance of the point-based method by introducing a novel glance and gaze manner. These methods are simpler, but are weak in modeling long range contextual information, resulting in poor performance when human and object are far apart.

\noindent	
{\bfseries Transformers-Based HOI Detection.} Recently, Transformers-based methods have been proposed to handle HOI detection as a set prediction problem. More specifically, \cite{qpic, endd} design the HOI instances as some learnable queries, and use a typical Transformer with encoder-decoder architectures to directly predict HOI instances in an end-to-end manner. In addition, \cite{hotr, chen2021reformulating} detect human-object instances and interaction labels with parallel decoders. In~\cite{zhang2021mining}, human/object detection and interaction classification are disentangled in a cascade manner. However, these methods adopt the simple tokenization strategy to structurize an image into a sequence of local patches, which is insufficient for detailed image understanding tasks like HOI detection.
	
\begin{figure}[t]
		\centering
		\includegraphics[width=0.99\linewidth]{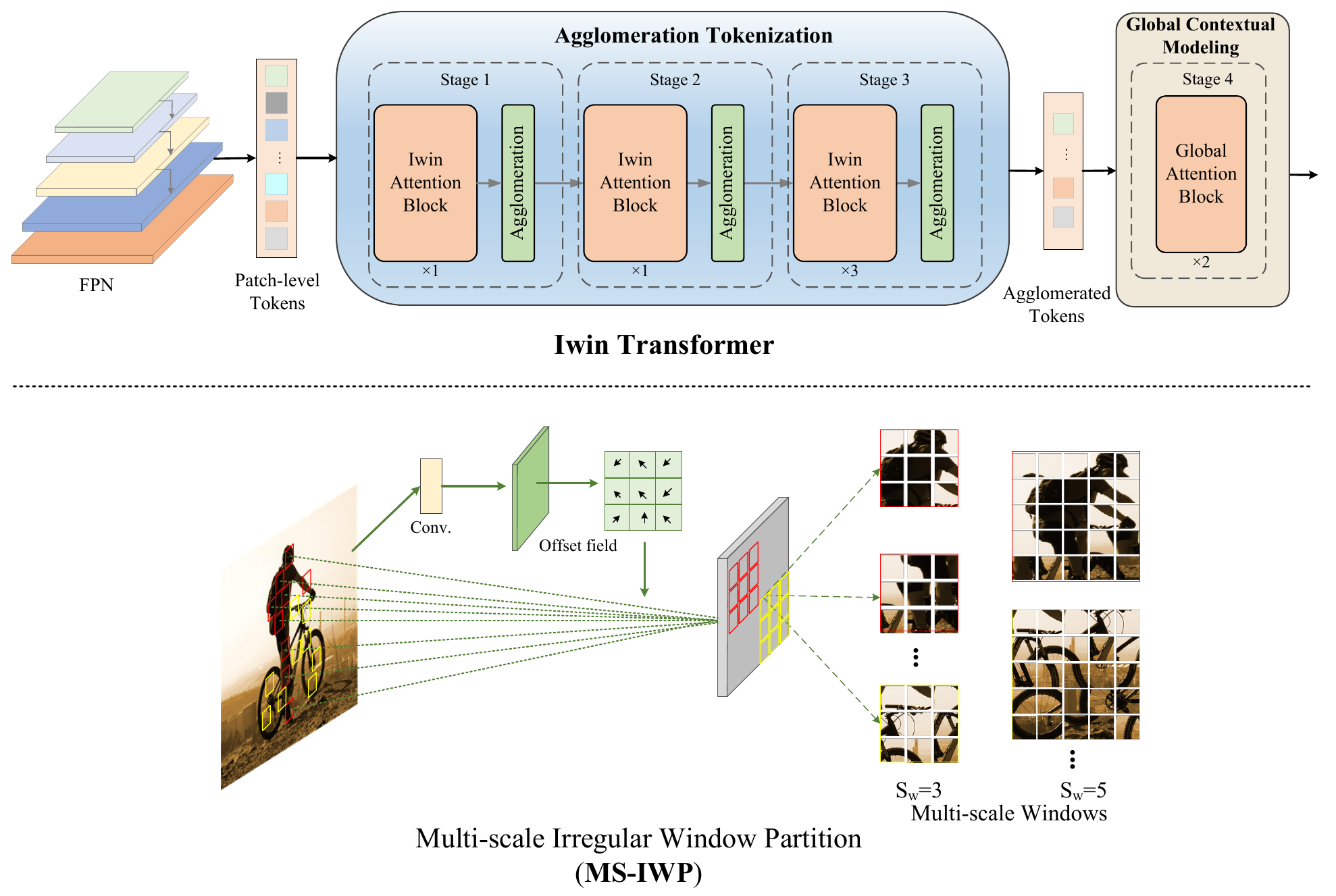}
		\caption{{\bfseries An overview of Iwin Transformer.} For simplicity of presentation, the decoder and matching component are omitted. The detailed illustrations about ``Iwin Attention Block'' and ``Agglomeration'' can be seen in Figure~\ref{whole}.} 
		\label{whole_view}
\end{figure}

\begin{figure}[t]
		\centering
		\includegraphics[width=0.95\linewidth]{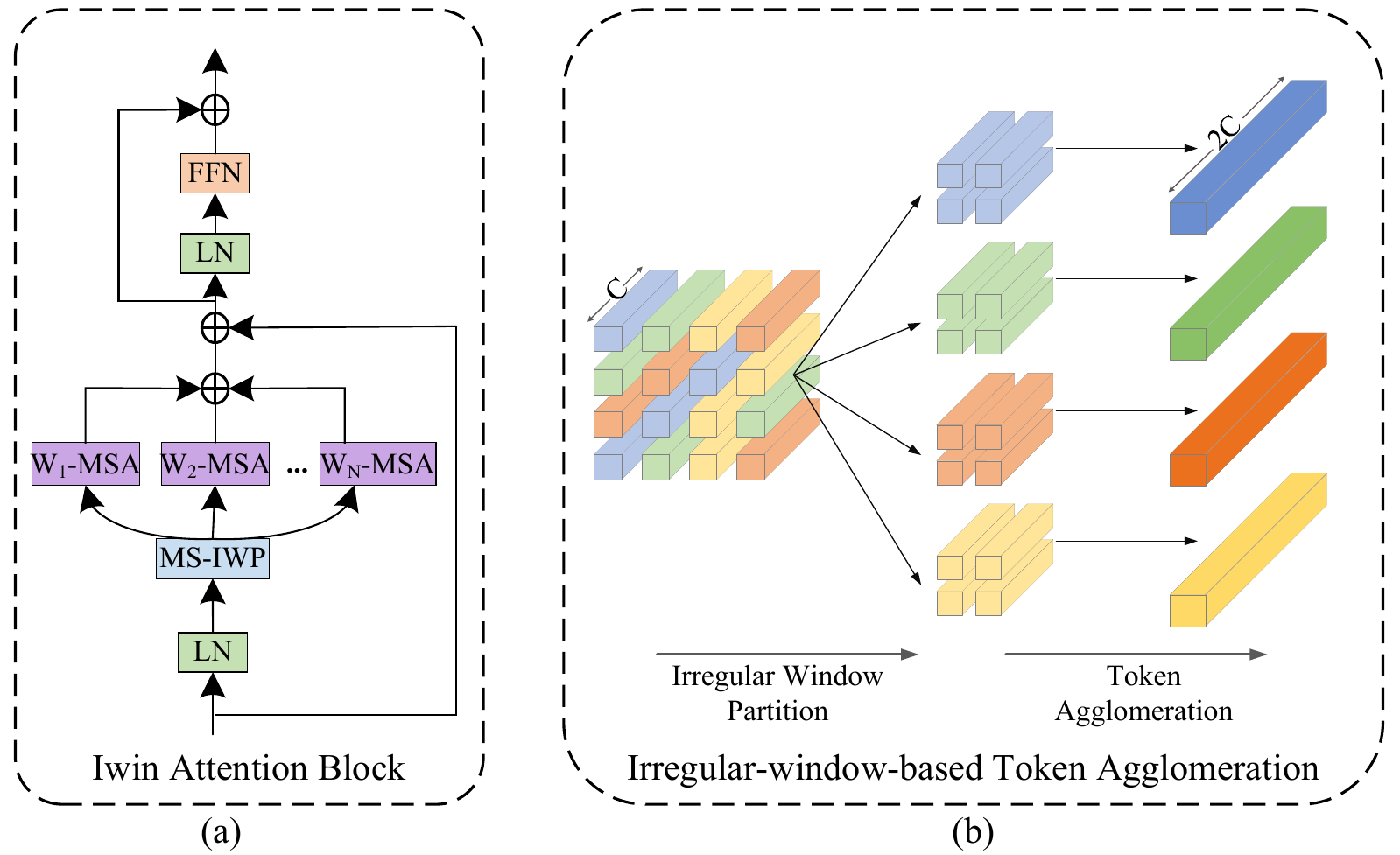}
		\caption{Iwin Attention Block \textbf{(a)} and Irregular Window-based Token Agglomeration \textbf{(b)}. ``LN'' refers to layer normalization.} 
		\label{whole}
\end{figure}

\section{Method}

An overview of the proposed Iwin Transformer is presented in Figure~\ref{whole_view}. In this section, we start with a general description of the entire model, followed by a detailed technical description of each key component.

\subsection{Architecture Overview}
{\bfseries Backbone.} Taking as input an image $\mathbf{x} \in \mathbb{R} ^{3 \times H \times W}$, the feature maps with different spatial resolutions are firstly computed by using a ResNet \cite{he2016deep} backbone to model local structures. Then, a feature pyramid network (FPN) is performed to get a high resolution feature map $\mathbf{z}_{b} \in \mathbb{R} ^{C_{b} \times \frac{H}{4} \times \frac{W}{4}}$ by weighted merging the feature maps with different resolutions, to ensure reliable performance in detecting small objects.Here, $C_b$ is the number of channels.

\noindent
{\bfseries Encoder} The encoder consists of two main parts: 1) agglomerative tokenization that progressively structurizes an image as a few agglomerated tokens by recursively performing irregular window-based token representation learning and agglomeration. 2) global contextual modeling for long-range context modeling.

\noindent
\emph{Agglomerative tokenization.} It consists of three ``Stage''s in a cascaded manner. Each stage is composed of token representation learning and token agglomeration within irregular windows. Specifically, token representation learning is performed by stacking several attention blocks based on multi-scale irregular window partition (MS-IWP), \emph{i.e.}, ``Iwin Attention Block'' in Figure~\ref{whole_view}. We describe it in detail in the next Section~\ref{irr-tpl}. 
After token representation learning, tokens in an irregular window are agglomerated into new tokens as the inputs for the next stage, which is described in detailed in Section~\ref{irr-ta}. 
By recursively performing irregular window-based token representation learning and agglomeration, an image can be structurized as a few agglomerated tokens with highly-abstracted semantics.

\noindent
\emph{Global contextual modeling.} After agglomerative tokenization, we then apply two cascaded global self-attention blocks to modeling the global contextual information. Unlike Iwin attention block, the attention weights are computed among all tokens in a global self-attention block. There are two main reasons: 1) interaction recognition demands a larger and more flexible receptive field, since the human and the object he/she interacts with may be far away or very close. 2) one interaction instance can provide some clues for the recognition of another interaction category, \emph{e.g.}, a man \underline{holding} a fork is likely to be \underline{eating} something. Although attention is  computed in a global manner, global self-attention block does not introduce too much computation since the number of tokens has been greatly reduced after agglomerative tokenization.\\

\noindent
{\bfseries Decoder.} The decoder consists of $6$ typical Transformer decoder layers, where each layer contains a self-attention module for correlations modeling between HOI instances, and a cross-attention module for HOI instances decoding. It takes as inputs both the outputs of encoder and $N$ learnable HOI queries with $256$ dimensions. We also conduct several different decoders, including multi-layer perceptron and local-to-global decoder which has a symmetrical structure of the proposed encoder. Their results are shown in Section~\ref{as}.

\noindent
{\bfseries Matching cost function.} We use the same matching cost function  proposed in \cite{qpic}. It is defined as:
\setlength{\abovedisplayskip}{2pt}
\setlength{\belowdisplayskip}{2pt}
\begin{equation}
\label{match}
\setlength{\abovedisplayskip}{2pt}
\setlength{\belowdisplayskip}{2pt}
\mathcal{L}_{cost} = \sum_{i}^{N} \mathcal{L}_{match}(\mathbf{g}^i, \mathbf{p}^{\sigma(i)}),
\end{equation}
where $ \mathcal{L}_{match}(\mathbf{g}^i, \mathbf{p}^{\sigma(i)}) $ is a matching cost between ground truth $\mathbf{g}^i$ and prediction $\mathbf{p}^{\sigma(i)}$. Specifically, the matching cost is designed as:
\begin{small}
	\begin{equation}\label{match-cost}
	\setlength{\abovedisplayskip}{2pt}
    \setlength{\belowdisplayskip}{2pt}
	\mathcal{L}_{match}(\mathbf{g}^i, \mathbf{p}^{\sigma(i)}) = \beta_1 \sum_{j \in h,o,r} \alpha_j \mathcal{L}_{cls}^{j} + \beta_2  \sum_{k \in h,o} \mathcal{L}_{box}^{k},
	\end{equation}
\end{small}
\noindent
where $\mathcal{L}_{cls}^{j} = \mathcal{L}_{cls}(g_j^i, p_j^{\theta(i)})$ is cross-entropy loss function, $j \in \{h,o,r\}$ denotes human, object or interaction, and $g_j^i$ represents the class label of $j$ on ground-truth $g^i$. $ \mathcal{L}_{box}^k$ is box regression loss for human box and object box, which is a weighted sum of GIoU \cite{giou} loss and $L_1$ loss. $\beta$ and $\alpha$ are both hyper-parameters. Finally, the Hungarian algorithm \cite{kuhn1955hungarian} is used to solve the following problem to find a bipartite matching:

\begin{equation}\label{bipar}
\setlength{\abovedisplayskip}{1pt}
\setlength{\belowdisplayskip}{2pt}
\hat{\sigma}=\underset{\sigma \in \mathfrak{S}_{N} }{\arg \min } \mathcal{L}_{\operatorname{cost}},
\end{equation}
where $\mathfrak{S}_{N}$ denotes the one-to-one matching solution space.

\subsection{Irregular Window Partition}
Window partition is firstly proposed in~\cite{swin}, which splits an image into several regular windows, as illustrated in Figure~\ref{ill}. Considering a feature map with resolution of $ H \times W $ and dimension of $C$, the computational complexity of a global multi head self-attention (G-MSA) module and a window-based one (W-MSA) with window size of $S_{w} \times S_{w} $ are:
\begin{align}
&\Omega (\text{G-MSA}) = 4HWC^2 + 2 (HW)^2C, \\
&\Omega (\text{W-MSA}) = 4HWC^2 + 2 S_{w}^2 HWC,
\end{align}
where the former is of quadratic computation \emph{w.r.t} pixel numbers while the latter is of linear computation. In regular window partition scheme, the tokens are sampled by using a regular rectangle $\mathcal{R} $ over the input feature map $ \mathbf{z}$, where the rectangle $\mathcal{R} $ defines the receptive field size. For example,
 $\mathcal{R} =  \{(0,0), (0, 1),...,(3, 4), (4, 4)\}$
defines a window with size of $5 \times 5$. However, as shown in Figure~\ref{ill}, regular windows may divide an object into parts and the tokens in a window are most likely unrelated, leading to redundancy in self-attention computing. Inspired by~\cite{dai2017deformable}, we propose irregular window partition by augmenting the regular windows with learned offsets $\mathbf{f_o}\in \mathbb{R} ^{2 \times H \times W}$. Specifically, the $\mathbf{f_o}$ is learned by performing a convolutional layer with a kernel size of $3 \times 3$ over the entire feature map. With the learned offsets $\mathbf{f_o}$, we can rearrange the tokens that are sampled from the irregular and offset locations $\mathbf{z}(\mathbf{p}_n + \Delta_{\mathbf{p}_n})$ to form a rectangular window, where $\mathbf{p}_n \in \mathcal{R}$ are the sampling locations for regular windows while $\Delta_{\mathbf{p}_n} \in \mathbf{f_o}$ are the learned offsets, $n=1,2,...,|\mathcal{R}|$. Since the learned offset $\Delta_{\mathbf{p}_n}$ is usually fractional, $\mathbf{z}(\mathbf{p})$ is defined via bilinear interpolation as
\begin{equation}\label{bil}
\mathbf{z}(\mathbf{p}) = \sum_{\mathbf{q}}K(\mathbf{q},\mathbf{p})\cdot \mathbf{z}(\mathbf{q}),
\end{equation}
where $\mathbf{p}$ denotes an arbitrary location $\mathbf{p}_n + \Delta_{\mathbf{p}_n}$, $\mathbf{q}$ enumerates all integral spatial locations, and $K(\cdot,\cdot)$ is the bilinear interpolation kernel. Since $K$ is two dimensional, it can be separated into two one dimensional kernels as
\begin{equation}\label{bil-fun}
K(\mathbf{p}, \mathbf{q}) = k(q_x, p_x) \cdot k(q_y, p_y),
\end{equation}
where $k(a,b) = \max(0, 1-|a-b|)$. 

  For convenience, we simply use the term ``an irregular window with size of $S_w \times S_w$'' to denote an irregular window that is generated by augmenting a regular rectangle with size of $S_w \times S_w$ in the following.

\subsection{Irregular-window-based Token Representation Learning}
\label{irr-tpl}

Irregular-window-based token representation learning is performed by stacked several Iwin attention blocks, where self-attention is computed within irregular windows. As illustrated in Figure~\ref{whole} (a), all Iwin attention blocks have an identical structure, which contains a multi-scale irregular window partition (MS-IWP) module, several window-based multi-head self-attention (W-MSA) modules as well as a feed-forward network (FFN). Specifically, in MS-IWP, a input feature map is dynamically splitted into several irregular windows by performing irregular window partitioning. Then, we rearrange the tokens within an irregular window to form a rectangular window for self-attention computition.
Moreover, the window size $S_w$ is designed to be various since the scales of the humans/objects in an image can be different, as illustrated in Figure~\ref{whole_view}. Specifically, $S_w \in \{5,7\}$ for ``Stage-1" and ``Stage-2" and $S_w \in \{3,5\}$ for ``Stage-3". Then, we apply multi-head self-attention within each window. As shown in Figure~\ref{whole}, window-based multi-head self-attention (W-MSA) is applied for $N$ times as there are windows of $N$ different sizes. After that, the output feature maps of different W-MSA modules are weighted summed via a convolutional layer with kernel size of $1 \times 1$. Finally, the FFN is applied to the sum of different W-MSA modules and input features, and each FFN is formed by two dense linear layers with ReLU \cite{relu} activations in between.

\subsection{Irregular-window-based Token Agglomeration}

\label{irr-ta}
As illustrated in Figure~\ref{whole} (b), we first perform irregular window partitioning with a size of $2 \times 2$, \emph{i.e.}, every 4 related tokens are grouped into one window. Then, these 4 tokens are concatenated to generate a $4C$-dimensional feature, where $C$ is the number of channels of one token. After that, we apply a linear layer on the $4C$-dimensional feature as 
\begin{equation}\label{merg}
\mathbf{t}_{new} = \mathbf{w} \cdot \text{concat}\left[\mathbf{z}(\mathbf{p}_n + \Delta_{\mathbf{p}_n}) | n \in 1,2,3,4\right],
\end{equation}
where $\mathbf{w}$ denotes the learned weights of linear layer. $\mathbf{p}_n$ and $\Delta_{\mathbf{p}_n}$ are regular grid locations and learned offsets, respectively. The output dimension of the linear layer is set as $2C$.

After each stage in tokenization, the number of tokens is reduced by a factor of $2 \times 2 =4$ , and the dimension of feature map are doubled (from $C$ to $2C$). Specifically, The number of tokens before and after `` Stage-$i$" are $\frac{H}{2^{1+i}} \times \frac{W}{2^{1+i}} $ and $\frac{H}{2^{2+i}} \times \frac{W}{2^{2+i}}$ respectively, and the output dimensions are $64 \times 2^{i-1}$. It not only further reduces the redundancy of tokens, but also enables the newly generated tokens to be aligned with humans/objects with different shapes and to characterize higher-abstracted visual semantics.

\section{Experiments}

\subsection{Datasets and Evaluation Metric}

\noindent
{\bfseries Datasets.} We conducted experiments on HICO-DET \cite{chao2018learning} and V-COCO \cite{gupta2015visual} benchmarks to evaluate the proposed method by following the standard scheme. Specifically, HICO-DET contains 38,118 and 9,658 images for training and testing, and includes 600 HOI categories (\emph{full}) over 117 interactions and 80 objects. It has been further split into 138 Rare (\emph{rare}) and 462 Non-Rare (\emph{non-rare}) HOI categories based on the number of training instances. V-COCO is a relatively smaller dataset that originates from the COCO \cite{coco}. It consists of 2,533 and 2,867 images for training, validation, as well as 4,946 ones for testing. The images are annotated with 80 object and 29 action classes.

\begin{small}
\begin{table*}[!t]
	\footnotesize
	\centering
	\resizebox{0.99\linewidth}{0.28\textheight}{
	\begin{tabular}{lcccccccccc}
		\toprule[1pt]
		\multirow{2}{*}[-0.5em]{Method} & \multirow{2}{*}[-0.5em]{Backbone} & \multirow{2}{*}[-0.5em]{Detector} & \multirow{2}{*}[-0.5em]{P} & \multirow{2}{*}[-0.5em]{L} & \multicolumn{3}{c}{Default} & \multicolumn{3}{c}{Known object} \\
		\cmidrule(lr){6-8}
		\cmidrule(lr){9-11}
		
		&  & & &  & Full$\uparrow$ & Rare$\uparrow$ & NonRare$\uparrow$ & Full$\uparrow$ & Rare$\uparrow$ & NonRare$\uparrow$ \\
		 \hline
		\multicolumn{2}{l}{\textit{CNN-based two-stage methods}}  & & & & & & & & & \\
		InteractNet~\cite{interactions}  & ResNet-50-FPN  & COCO  &  &               & 9.94   & 7.16  & 10.77  & -      & -      &  -             \\
		GPNN~\cite{qi2018learning}       & ResNet-101     & COCO  &  &               & 13.11  & 9.34  & 14.23  & -      & -      &  -              \\
		iCAN~\cite{ican}                 & ResNet-50      & COCO  &  &               & 14.84  & 10.45 & 16.15  & 16.26  & 11.33  & 17.73            \\
		No-Frills~\cite{no-frills}       & ResNet-152     & COCO  &  & $\checkmark$  & 17.18  & 12.17 & 18.68  & -      & -      &  -                \\
		TIN~\cite{li2019transferable}    & ResNet-50      & COCO  & $\checkmark$ &   & 17.22  & 13.51 & 18.32  & 19.38  & 15.38  & 20.57              \\
		PMFNet~\cite{wan2019pose}        & ResNet-50-FPN  & COCO  & $\checkmark$ &   & 17.46  & 15.65 & 18.00  & 20.34  & 17.47  & 21.20               \\
		CHG~\cite{wang2020contextual}    & ResNet-50      & COCO  &  &               & 17.57  & 16.85 & 17.78  & 21.00  & 20.74  & 21.08                \\
Peyre et al.~\cite{peyre2019detecting}   & ResNet-50-FPN  & COCO  &  & $\checkmark$  & 19.40  & 14.63 & 20.87  & -      & -      &  -                    \\
		VSGNet~\cite{ulutan2020vsgnet}   & ResNet152      & COCO  &  &               & 19.80  & 16.05 & 20.91  & -      & -      &  -                     \\
		FCMNet~\cite{liu2020amplifying}  & ResNet-50      & COCO  & $\checkmark$ & $\checkmark$  & 20.41  & 17.34  & 21.56  & 22.04   &  18.97  & 23.12    \\
		ACP~\cite{kim2020detecting}      & ResNet-152     & COCO  & $\checkmark$ & $\checkmark$  & 20.59  & 15.92  & 21.98  & -      & -      &  -          \\
		PD-Net~\cite{zhong2021polysemy}  & ResNet-152     & COCO  &  & $\checkmark$  & 20.81  & 15.90 & 22.28  & 24.78  & 18.88  & 26.54                     \\
		PastaNet~\cite{li2020pastanet}   & ResNet-50      & COCO  & $\checkmark$ & $\checkmark$  & 22.65  & 21.17  & 23.09  & 24.53   &  23.00  & 24.99       \\
		VCL~\cite{zhi_eccv2020}          & ResNet101      & COCO  &  &               & 19.43  & 16.55 & 20.29  & 22.00  & 19.09  &  22.87                      \\
		DRG~\cite{drg}                   & ResNet-50-FPN  & COCO  &  & $\checkmark$  & 19.26  & 17.74 & 19.71  & 23.40  & 21.75  &  23.89                       \\ 
 Zhang et al. ~\cite{zhang2021spatially} & ResNet-50-FPN  & COCO  &  &               & 21.85  & 18.11 & 22.97  & -      & -      &  -                            \\
		
		\midrule[1pt]
		
		\multicolumn{2}{l}{\textit{CNN-based single-stage methods}} & & & & & & & & & \\
		
		UnionDet~\cite{uniondet}         & ResNet-50-FPN  & HICO-DET   &  &  & 17.58  & 11.52  & 19.33  & 19.76  & 14.68  & 21.27              \\ 
		IPNet~\cite{ipnet}               & Hourglass      & COCO       &  &  & 19.56  & 12.79  & 21.58  & 22.05  & 15.77  & 23.92               \\
		PPDM~\cite{ppdm}                 & Hourglass      & HICO-DET   &  &  & 21.73  & 13.78  & 24.10  &  24.58 & 16.65  & 26.84                \\
		GGNet~\cite{ggnet}               & Hourglass      & HICO-DET   &  &  & 23.47  & 16.48  & 25.60  & 27.36  & 20.23  & 29.48                 \\
		ATL ~\cite{hou2021affordance}    & ResNet-50      & HICO-DET   &  &  & 23.81  & 17.43  & 25.72  & 27.38  & 22.09  & 28.96                  \\
		\midrule[1pt]
		
		\multicolumn{2}{l}{\textit{Transformer-based methods}}  & & & & & & & & & \\
		
		HOI-Trans~\cite{endd}             & ResNet-50      & HICO-DET   &  &  & 23.46  & 16.91  & 25.41  & 26.15  & 19.24  & 28.22                   \\
		HOTR~\cite{hotr}                 & ResNet-50      & HICO-DET   &  &  & 25.10  & 17.34  & 27.42  & -      & -      & -                        \\
	AS-Net~\cite{chen2021reformulating}  & ResNet-50      & HICO-DET   &  &  & 28.87  & 24.25  & 30.25  & 31.74  & 27.07  & 33.14                     \\
		QPIC~\cite{qpic}                 & ResNet-50      & HICO-DET   &  &  & 29.07  & 21.85  & 31.23  & 31.68  & 24.14  & 33.93                      \\
		\rowcolor{cyan!50}
		Iwin-S (Ours)                    & ResNet-50-FPN  & HICO-DET   &  &  & 24.33  & 18.50  & 26.04  & 28.41  & 20.67  & 30.17                       \\
		\rowcolor{cyan!50}
		Iwin-B (Ours)                    & ResNet-50-FPN  & HICO-DET   &  &  & \textbf{\color{blue}{32.03}}  & \textbf{\color{blue}{27.62}}  & \textbf{\color{blue}{34.14}}  & \textbf{\color{blue}{35.17}}  & \textbf{\color{red}{28.79}}  & \textbf{\color{blue}{35.91}}                        \\
		\rowcolor{cyan!50}
		Iwin-L (Ours)                    & ResNet-101-FPN & HICO-DET   &  &  & \textbf{\color{red}{32.79}}  & \textbf{\color{red}{27.84}}  & \textbf{\color{red}{35.40}}  & \textbf{\color{red}{35.84}}  & \textbf{\color{blue}{28.74}}  & \textbf{\color{red}{36.09}}                         \\
		
	\bottomrule[1pt]
		
	\end{tabular}
	}
	\caption{{\bfseries Performance comparison on the HICO-Det test set.} `P' refers to human pose and `L' denotes language. The Best performance are represented in \textbf{\color{red}{red}} and the second best ones are shown in \textbf{\color{blue}{blue}.}}
	\label{tab:hico}
\end{table*}
\end{small}
\noindent
{\bfseries Evaluation metric.} We use the commonly used mean average precision (mAP) to evaluate model performance on both datasets. A predicted HOI instance is considered as true positive if and only if the predicted human and object bounding boxes both have IoUs larger than 0.5 with the corresponding ground-truth bounding boxes, and the predicted action label is correct. 

Moreover, for HICO-DET, we evaluate model performance in two different settings following \cite{chao2018learning}: (1) \emph{Known-object setting}. For each HOI category, we evaluate the detection only on the images containing the target object category. (2) \emph{Default setting}. For each HOI category, we evalute the detection on the full test set, including images that may not contain the target object.

\subsection{Implementation Details}
We implemented three variant architectures of Iwin: Iwin-S, Iwin-B, and Iwin-L, where ``S'', ``B'' and ``L'' refer to small, base and large, respectively. The number of blocks in these model variants are:
\begin{itemize}
	\setlength{\itemsep}{3pt}
    \setlength{\parsep}{3pt}
    \setlength{\parskip}{3pt}
	\item Iwin-S: $D_c=32$, block numbers = $\{1,1,1,1\},$
	\item Iwin-B: $D_c=64$, block numbers = $\{1,1,3,2\},$
	\item Iwin-L: $D_c=64$, block numbers = $\{1,1,3,2\},$
\end{itemize}
where we applied ResNet-50 as a backbone feature extractor for both Iwin-S and Iwin-B, and ResNet-101 was utilized for Iwin-L. Besides, the decoder contains 6 Transformers decoder layers and the hyper-parameters $\beta_1$, $\alpha$ and $\beta_2$ in the loss function are set as 1, 1, 2.5 for all experiments. The number of queries $N$ was set to 50 for HICO-DET and 100 for V-COCO.
\begin{table}[t]
    \begin{minipage}[c]{0.49\textwidth}
    \centering
	\small
	\centering
	\scalebox{0.8}{
	\begin{tabular}{lccccc}
		\toprule
		Method & Backbone & P & L & $AP_{role}$ \\
		
		 \hline
		\multicolumn{2}{l}{\textit{CNN-based two-stage methods}}  & & &   \\
		InteractNet~\cite{interactions}  & ResNet-50-FPN  &  &               & 40.00            \\
		GPNN~\cite{qi2018learning}       & ResNet-101     &  &               & 44.00             \\
		iCAN~\cite{ican}                 & ResNet-50      &  &               & 45.30              \\
		TIN~\cite{li2019transferable}    & ResNet-50      &  &               & 47.80               \\
		VSGNet~\cite{ulutan2020vsgnet}   & ResNet152      &  &               & 51.80                \\
		PMFNet~\cite{wan2019pose}        & ResNet-50-FPN  & $\checkmark$ &   & 52.00                 \\
		CHG~\cite{wang2020contextual}    & ResNet-50      &  &               & 52.70                  \\
		FCMNet~\cite{liu2020amplifying}  & ResNet-50      & $\checkmark$ & $\checkmark$  & 53.10       \\
		ACP~\cite{kim2020detecting}      & ResNet-152     &  & $\checkmark$  & 53.23                   \\
		
		\midrule[1pt]
		
		\multicolumn{2}{l}{\textit{CNN-based single-stage methods}} & & & \\
		
		UnionDet~\cite{uniondet}         & ResNet-50-FPN  &  &  & 47.50              \\ 
		IPNet~\cite{ipnet}               & Hourglass      &  &  & 51.00               \\
		GGNet~\cite{ggnet}               & Hourglass      &  &  & 54.70                 \\
		\midrule[1pt]
		
		\multicolumn{2}{l}{\textit{Transformer-based methods}}  & & &\\
		
		HOI-Trans~\cite{endd}             & ResNet-50      &  &  & 52.90                   \\
		HOTR~\cite{hotr}                 & ResNet-50      &  &  & 55.20                    \\
	AS-Net~\cite{chen2021reformulating}  & ResNet-50      &  &  & 53.90                     \\
		QPIC~\cite{qpic}                 & ResNet-50      &  &  & 58.80                      \\
		\rowcolor{cyan!50}
		Iwin-S (Ours)                    & ResNet-50-FPN  &  &  & 51.81                       \\
		\rowcolor{cyan!50}
		Iwin-B (Ours)                    & ResNet-50-FPN  &  &  & \textbf{\color{blue}{60.47}}                        \\
		\rowcolor{cyan!50}
		Iwin-L (Ours)                    & ResNet-101-FPN &  &  & \textbf{\color{red}{60.85}}                         \\
		
	\bottomrule[1pt]
		
	\end{tabular}
	}
	\caption{ Performance comparison on the V-COCO test set.}
	\label{tab:vcoco}

    \end{minipage}
    \begin{minipage}[c]{0.5\textwidth}

    \centering
	\small
	\centering
	\scalebox{0.84}{
	\begin{tabular}{ccccc|cccc}
		\toprule[1pt]
		FPN & IWP        & MS              & TA           & GC         & Full$\uparrow$  & Rare$\uparrow$  & NoneRare$\uparrow$  \\
		\midrule
		                & $\checkmark$    & $\checkmark$ & $\checkmark$   & $\checkmark$  & 31.24           & 26.01           & 32.95                \\
		 $\checkmark$   &                 & $\checkmark$ & $\checkmark$   & $\checkmark$  & 29.12           & 24.74           & 31.08                 \\
		 $\checkmark$   & $\checkmark$    &              & $\checkmark$   & $\checkmark$  & 31.42           & 26.97           & 33.15                  \\
		 $\checkmark$   &                 &              & $\checkmark$   & $\checkmark$  & 26.33           & 24.05           & 28.21                  \\
		 $\checkmark$   & $\checkmark$    & $\checkmark$ &                & $\checkmark$  & 30.16           & 25.68           & 32.49                   \\
		 $\checkmark$   & $\checkmark$    & $\checkmark$ & $\checkmark$   &               & 23.05           & 16.50           & 24.62                    \\
		 \rowcolor{cyan!50}
		 $\checkmark$   & $\checkmark$    & $\checkmark$ & $\checkmark$   & $\checkmark$  & \textbf{32.03}  & \textbf{27.62}  & \textbf{34.14}            \\
		
	\bottomrule[1pt]
		
	\end{tabular}
	}
	\caption{{\bfseries The effects of different modules}. ``IWP'': irregular window partition, ``MS'': multi-scale windows, ``TA'': token agglomeration and ``GC'': global contextual modeling.}

	\label{tab:components}

	\small
	\centering
	\scalebox{0.92}{
	\begin{tabular}{c|ccc}
		\toprule[1pt]
		Strategy         & Full$\uparrow$  & Rare$\uparrow$  & NoneRare$\uparrow$  \\
		\midrule
		 Norm-window              & 27.50           & 24.27           & 29.84                \\
		 K-means                   & 28.19           & 24.72           & 30.12                 \\
		 \rowcolor{cyan!50}
		 Irregular-window        & \textbf{32.03}  & \textbf{27.62}  & \textbf{34.14}         \\		
	\bottomrule[1pt]
		
	\end{tabular}
	}
	\caption{Comparison between different types of token agglomeration strategy.}
	\label{tab:merging_strategy}
    \end{minipage}

\end{table}
Unlike existing methods that initialized the network by the parameters of DETR trained with the COCO dataset, which contains prior-information about objects, we train Iwin Transformer from scratch. We employed an AdamW \cite{adamw} optimizer for 150 epochs as well as a multi-step decay learning rate scheduler. A batch size of 16, an initial learning rate of 2.5e-4 for Transformers and 1e-5 for backbone are used. The learning rate decayed by half at 50, 90, and 120 steps respectively.

\subsection{Comparisons with State-of-the-Art}

We first summarize the main quantitative results in terms of mAP on HICO-DET in Table \ref{tab:hico} and $AP_{role}$ on V-COCO in Table \ref{tab:vcoco}.
As shown by the results, Transformers-based methods show great potential compared to the CNN-based methods. This can mainly attribute to the ability of self-attention to selectively capture long range dependence, which is essential for HOI detection. On the basis of that, our method outperforms SOTA approaches. Specifically, Iwin achieves 3.7 mAP gain compared with QPIC \cite{qpic} on \emph{Full} setting of HICO-DET dataset as well as  2.0 $AP_{role}$ gain on V-COCO. The main reason for such results is that we conduct token representation learning and agglomeration in a more effective manner. Firstly, instead of utilizing global attention, we novelly perform self-attention in irregular windows for token representation learning. It allows Iwin effectively eliminates redundancy in self-attention while achieves a linear computational complexity. Secondly, Iwin introduces a new irregular window-based token agglomeration strategy, which progressively structurizes an image as a few agglomerated tokens with highly-abstracted visual semantic, which enables the contextual relations to be more easily captured for interaction recognition. Moreover, Iwin can leverage high-resolution feature maps due to its computational efficiency, which also boosts the performance in detecting small objects.

\subsection{Ablation Study}

\label{as}
\subsubsection{Model components}\

\noindent
In this subsection, we analyze the effectiveness of the proposed strategies and components in detail. All experiments are performed on the HICO-DET dataset and the results are reported under the \emph{Default} setting. Due to space limitations, some other important experiments can be seen in the \emph{supplementary material}. 

\noindent
{\bfseries FPN.} Feature pyramid network~\cite{lin2017feature} was first proposed to solve the problem of detecting small objects. It is also crucial for HOI detection since the objects with which a human interacts may be small, such as a cup, a spoon and a mobile phone. It has been applied in lots of CNN-based HOI detection methods \cite{interactions, wan2019pose, drg}. However, limited by the high computational complexity, existing Transformers-based HOI detection methods can not process such high resolution feature maps, resulting in poor performance in detecting small objects. In contrast, Iwin is friendly to high resolution feature maps thanks to the strategies of irregular window-based attention and token agglomeration. As shown in Table \ref{tab:components}, with FPN, the model obtains 0.79 mAP gain.

\begin{table}[t]
    \begin{minipage}[c]{0.51\textwidth}
    \centering
    
	\small
	\setlength{\tabcolsep}{2pt}
	\centering
	\scalebox{0.9}{
	\begin{tabular}{c|ccc}
		\toprule[1pt]
		Position Encoding         & Full$\uparrow$  & Rare$\uparrow$  & NoneRare$\uparrow$  \\
		\midrule
		 None                     & 20.80           & 17.54           & 22.30                \\
		 Sin at input            & 28.74           & 25.83           & 31.07                 \\
		 Sin for all             & 28.59           & 25.74           & 30.86                   \\
		 Learned at attn.         & 31.64           & 27.03           & 32.79                    \\	
		 \rowcolor{cyan!50}
		 Sine at attn.            & \textbf{32.03}  & \textbf{27.62}  & \textbf{34.14}         
		                \\
	\bottomrule[1pt]
		
	\end{tabular}
	}
	\caption{ The effects of different position encoding strategies.}
	\label{tab:position}

    \end{minipage}
    \begin{minipage}[c]{0.49\textwidth}

    \centering
    
	\small
	\setlength{\tabcolsep}{3pt}
	\centering
	\scalebox{0.9}{
	\begin{tabular}{c|ccc}
		\toprule[1pt]
		Decoder         & Full$\uparrow$  & Rare$\uparrow$  & NoneRare$\uparrow$  \\
		\midrule
		 MLP            & 19.39           & 16.41           & 21.85                \\
		 Symmetry       & 26.03           & 24.11           & 29.02                 \\
		Trans$\times 2$ & 27.31           & 25.07           & 30.28                  \\
		Trans$\times 4$ & 29.40           & 26.16           & 31.46                   \\
		\rowcolor{cyan!50}
		Trans$\times 6$ & \textbf{32.03}  & \textbf{27.62}  & \textbf{34.14}                   \\		
	\bottomrule[1pt]
		
	\end{tabular}
	}
	\caption{ The effects of different types of decoder.}

	\label{tab:decoder}

    \end{minipage}

\end{table}

\noindent
{\bfseries Multi-scale irregular window partition.} There are two main strategies in a MS-IWP module, including irregular window partition (IWP) and multi-scale windows (MS). Augmented by learned offsets, IWP enables the irregular windows to align with human/object with different shapes and to eliminate redundancy in token representation learning. Moreover, it allows the irregular windows to own a dynamic receptive field compared to the regular ones, leading to efficient human/object detection. As shown in Table \ref{tab:components}, it plays an essential role in IWP and achieves 3.1 mAP gain. Besides, we encourage the window size to be diverse so that they can handle different sizes of objects. When we use neither of the two strategies, the performance of the model is severely degraded, from 32.03 to 26.33 mAP, which indicates that learning strong token representation is important for HOI detection.

\begin{figure}[!t]

	\centering
	\includegraphics[width=0.99\linewidth]{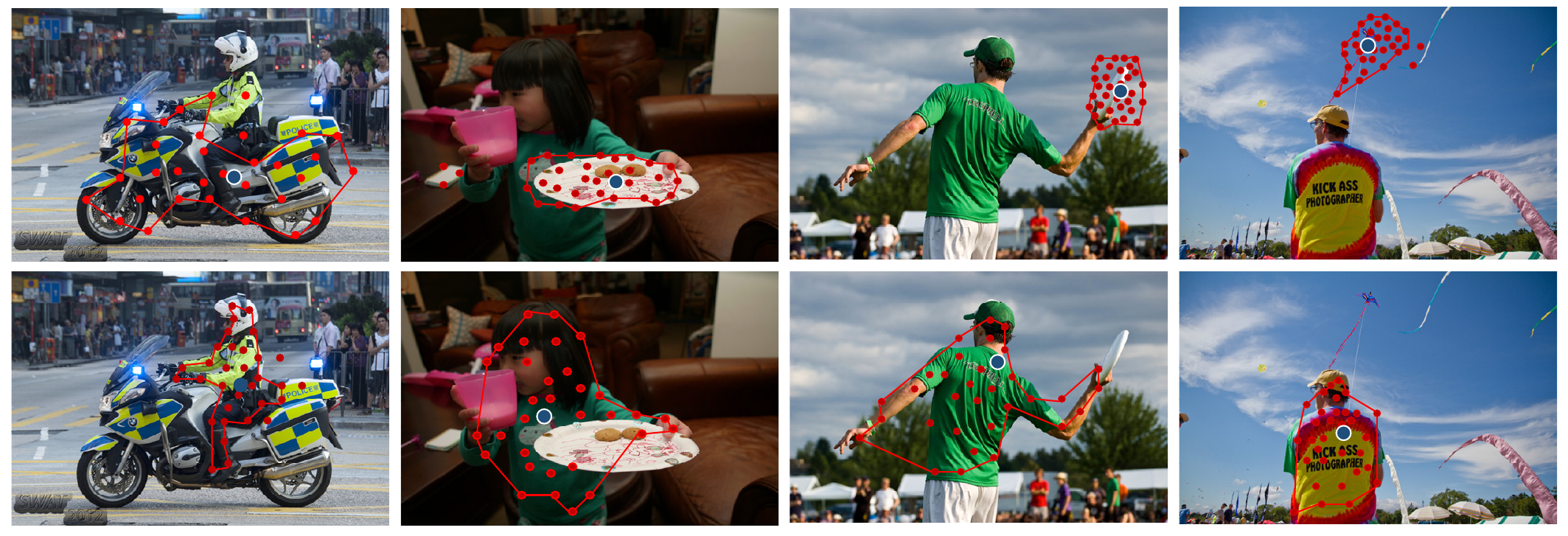}
	\caption{{\bfseries Visualization of irregular windows}.  Each \textbf{\color{blue}{blue}} point is the location of an agglomerated token obtained by agglomerative tokenization, and the \textbf{\color{red}{red}} points are the locations of the tokens involved in agglomeration. The envelope line of the \textbf{\color{red}{red}} points shows the shape of the irregular window for the agglomerated token. As the input of Iwin Transformer are feature maps outputted by the backbone, the locations in the original image are calculated via bilinear interpolation. Best view in color.}
	\label{merging}

\end{figure}

\noindent
{\bfseries Token agglomeration.}  Irregular window-based token agglomeration progressively structurizes an image as few agglomerated tokens with highly-abstracted visual semantics. It gradually reduce the number of tokens, ensures the self-attention to be performed in a global manner. Specifically, we illustrate the learned sampling locations in Figure~\ref{merging}. There are totally $4^3=64$ sampling points for each of the agglomerated token obtained by agglomerative tokenization, since every $4$ tokens are fused in a merging layer and $3$ layers are employed. It can be seen from the figure that the envelope line of the sampling locations shows the shape of the irregular window, which is aligned with a semantic region with different shapes. In addition, we have performed several other merging strategies and list the quantitative results in Table \ref{tab:merging_strategy}. The ``Norm-Window'' refers to fusing $4$ neighbouring tokens in a regular window, \emph{i.e.}, a rectangle with size of $2 \times 2$.  ``K-means'' denotes using an additional k-means algorithm~\cite{macqueen1967some} to cluster the similar tokens. For the former, it is more of a regular downsampling operation rather than generating high-level object-level semantic information, where the performance of model is reduced to $27.5$.  Meanwhile, when applying k-means algorithm, it is hard to determine the value of $K$, and as an individual module, it has to be trained separately.

\begin{figure}[t]
    \centering
    \begin{minipage}[t]{0.49\textwidth}
    \centering
    \includegraphics[width=0.99\linewidth]{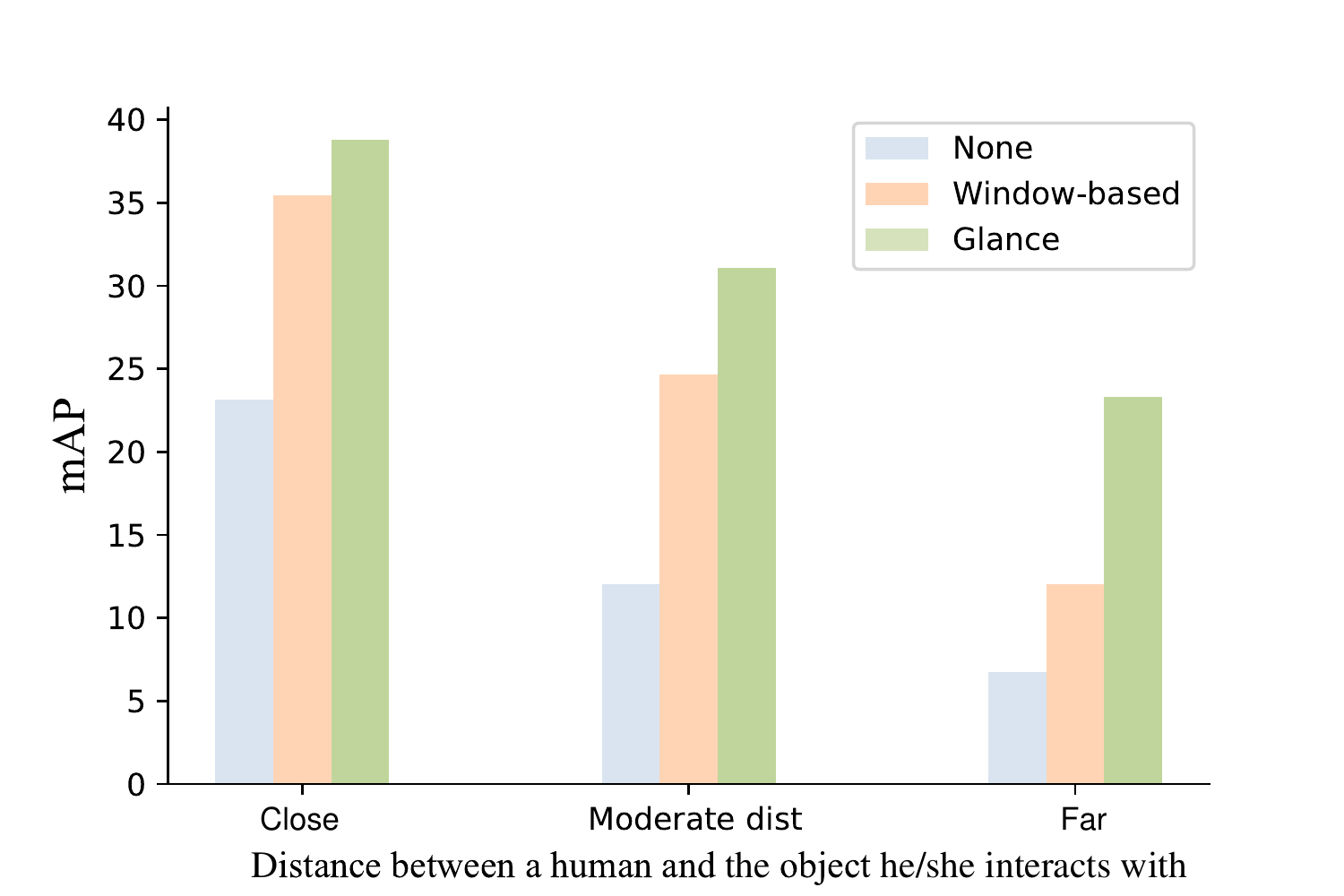}
    \caption{ Performance of different types \textbf{global contextual modeling} on different spatial distributions. The mAP is reported under the \emph{Full} setting.}
    \label{dist}
    \end{minipage}
    \begin{minipage}[t]{0.49\textwidth}
    \centering
    \includegraphics[width=0.99\linewidth]{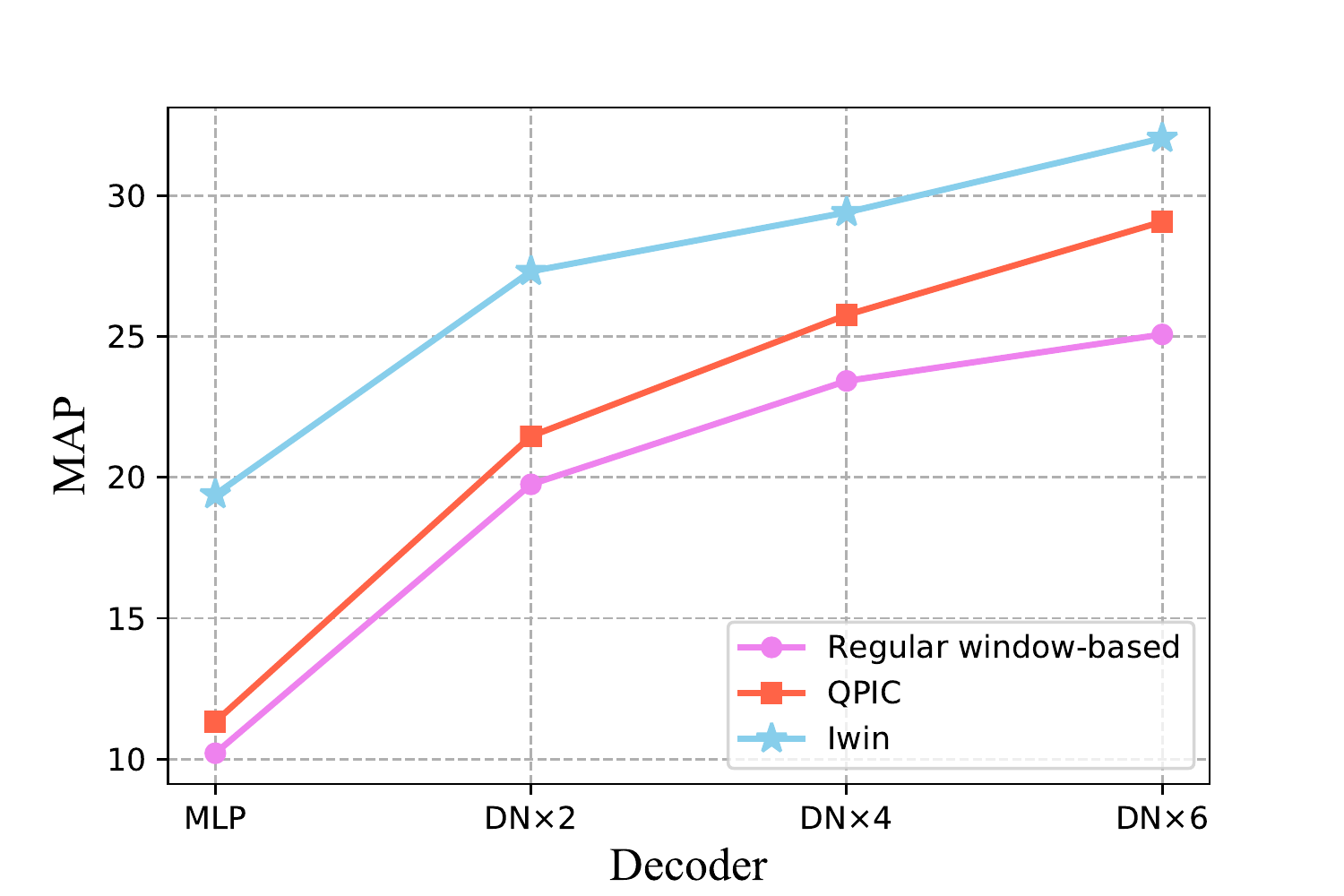}
    \caption{ Performance of different types of decoder. ``DN $\times M$'' refers to stacking $M$ typical Transformer decoder layers. The mAP is reported under the \emph{Full} setting.}
    \label{decoders}
    \end{minipage}

    \end{figure}

\noindent
{\bfseries Global contextual modeling.}
Global contextual modeling consists of two global self-attention mechanisms. As an essential component, it achieves a gain of 9+ mAP. To further validate the importance of global self-attention for interaction recognition, we perform the global contextual modeling in two extra different manners: remove this module directly and replace the global attention with window-based attention. Their results are shown in Figure \ref{dist}. Specifically, on the basis of normalizing both the height and width of image, we divide the $L_1$ distance between a human center and an object center into three uniform parts and denote them as ``close'', ``moderate dist.'' and ``far'', respectively. Without global contextual modeling, the model has unsatisfactory performance in all cases. When human and object are close, window-based attention achieves similar performance with global attention. However, as the distance increases, the performance of window-based attention is degraded seriously while the global attention can still work well. The main reason is that the ROI of interaction recognition can be diversely distributed image-wide.

\noindent
{\bfseries Position encoding.} Position encoding is essential for Transformer architecture since it is permutation invariant. However, in Iwin, convolutional operation is also employed to learn the offsets in irregular windowing, in which position encoding is not a necessity. Therefore, we conduct more experiments to explore the effect of position encoding on convolution operation and the results are shown in Table~\ref{tab:position}. Specifically, ``Sin'' refers to using fixed sinusoidal functions and ``Learned'' denotes using learning embeddings. ``At input'' refers the position encoding is utilized for only one time that directly be added to the output features of backbone. ``At attention'' implies the position encoding is used only when computing attention weights and ``For all'' indicts the position encodings are employed in both attention modules and convolution operations. As the result shown, position encoding is important for attention computing but of little significant for convolution ones.

\noindent
{\bfseries Decoder.} We simply use the typical Transformers decoder for our approach. As shown in Table \ref{tab:decoder}, when we replace the Transformers decoder with a simple multi-layer perceptron (MLP), the model still has a competitive performance to some recent CNN-based methods. It suggests that our encoder has the ability to effectively transform the visual information into high-level semantic space of HOI. Besides, we also design a decoder that has a symmetry structure with the encoder, where attention is conducted in both local and global manners. However, it does not achieve the desired effect. There are two different types of attention in decoder: self-attention and cross-attention. Specifically, self-attention is responsible for learning the correlation among the HOI queries. For example, a man holding a fork is more likely to be eating. Meanwhile, with self-attention, the HOI instances are decoded from the high-level semantic space generated by the encoder. As these two both require a large receptive field, it is more effective to perform the attention in a global manner. Besides, the model performs better with more decoder layers.

\subsubsection{Importance of irregular windowing}\

As mentioned above, HOI detection is a sequential process of human/object detection and interaction recognition. Unlike object detection that can be performed by modeling local structures, interaction recognition demands a larger receptive field and higher-level semantic understanding. To verify the capacity of different models to extract highly-abstracted features, we employ several different types of decoder, from simple to cpmplex, the results are illustrated in Figure~\ref{decoders}. The relative gaps of performance between Iwin and the other methods become more evident as the decoder gets simper. There are two main reasons: 1) computing attention weights among all tokens introduces severe redundancy in self-attention for QPIC~\cite{qpic}. Regular window-based Transformer has the same weakness since the tokens in a regular window are most likely to be irrelevant. 2) without token agglomeration, QPIC-liked methods deal with local patch-based tokens, which can only model the local structures needed for object detection. With irregular window-based token representation learning and agglomeration, Iwin's robust highly-abstracted visual semantic extraction capability are shown in two aspects. Firstly, even using a quite simple decoder, \emph{e.g.}, a MLP consists of two dense-connected layers, Iwins achieves competitive performance compared to some recent CNN-based methods. Secondly, Iwin is trained from scratch to get the SOTA results while most existing methods utilize pre-trained models, such as DETR~\cite{detr} and CLIP~\cite{clip}. Besides, unlike existing methods, Iwin is easy to train by performing irregular window partition. As shown in Figure~\ref{first}, when training from scratch, Iwin only needs half of the training epoches compared to other methods.

\section{Conclusion}

In this paper, we propose Iwin transformer, a novel vision Transformer for HOI detection, which progressively performs token representation learning and token agglomeration within irregular windows. By employing irregular window partition, Iwin Transformer can eliminate redundancy in token representation and generate new tokens to align with humans/objects with different shapes, enables the extraction of highly-abstracted visual semantics for HOI detection. We validate Iwin transformer on two challenging HOI benchmarks and achieve considerable performance boost over SOTA results.

\clearpage
%
%
\bibliographystyle{splncs04}
\bibliography{egbib}
\end{document}